%% file: main.tex
\documentclass{article}

\usepackage{arxiv}

\usepackage[utf8]{inputenc}
\usepackage[T1]{fontenc}
\usepackage{hyperref}
\usepackage{url}
\usepackage{booktabs}
\usepackage{amsfonts}
\usepackage{nicefrac}
\usepackage{microtype}
\usepackage{cleveref}
\usepackage{graphicx}
\usepackage[numbers,square]{natbib}
\usepackage{doi}

\graphicspath{{figures/}}

\title{CrabOS: An Operating System for Human-AI Co-inhabitation}
\input{author}
\date{}

\renewcommand{\shorttitle}{CrabOS}

\hypersetup{
  pdftitle={CrabOS: An Operating System for Human-AI Co-inhabitation},
  pdfauthor={Author One, Author Two},
  pdfkeywords={human-AI co-inhabitation, agent system, operating system},
}

\begin{document}
\maketitle

\begin{abstract}
\input{sections/00_abstract}
\end{abstract}

\keywords{human-AI co-inhabitation \and agent system \and operating system}

\input{sections/01_introduction}
\input{sections/02_related_work}
\input{sections/03_design_principles}
\input{sections/04_system_design}
\input{sections/05_implementation}
\input{sections/06_case_studies}
\input{sections/07_discussion}
\input{sections/08_conclusion}

\bibliographystyle{unsrtnat}
\bibliography{references}

\end{document}

%% file: author.tex
\author{
  Qi Yang \\
  \texttt{qi.yang@stu.pku.edu.cn} \\
  Institute for Artificial Intelligence, Peking University
  \And
  Yun Ma \\
  \texttt{mayun@pku.edu.cn} \\
  Institute for Artificial Intelligence, Peking University
}

%% file: sections/00_abstract.tex
AI agents are evolving into long-running computational entities that can invoke tools, maintain memory, and complete complex tasks across applications. In real-world settings, completing a task often requires humans and AI to take turns leading its execution. Such alternation depends on the seamless handoff of the work state of the task between humans and AI. Existing agent systems, however, provide humans and AI with separate work environments. AI agents must therefore rely on additional bridges to continue work: either developers build task-specific interfaces to access the work state, or users manually transfer relevant parts of it through screenshots or textual descriptions. Both approaches make handoffs costly and scale poorly.

We propose \textit{Human-AI Co-inhabitation}, a type of work environment that enables humans and AI to seamlessly take turns continuing work on the same task, and design and implement CrabOS to realize this concept. CrabOS\footnote{Project page: https://openbaylab.github.io/CrabOS/} represents the work state as natural-language-readable text objects shared by humans and AI, allowing both to access and manipulate it directly through the same auditable interface without bridges. Case studies show that CrabOS elevates support for complex tasks with alternating human and AI leadership from bridge-dependent application-level solutions to native operating-system capabilities, which provide a new foundation for developing and running AI agents.

%% file: sections/01_introduction.tex
\section{Introduction}
\label{sec:introduction}

As large language models advance, AI agents are evolving from conversational tools that respond passively into computational entities that can autonomously invoke tools, maintain memory, and work continuously on complex tasks. AI thus becomes, like a human, an executor able to independently move a task forward. Complex tasks, however, are rarely completed by the same executor from start to finish. Instead, humans and AI take turns moving the task forward through successive stages. After each stage, responsibility for the task passes to the next executor. Consider a user conducting research with an AI agent. After running an experiment, the agent leaves substantial temporary data and intermediate results that the user wants to visualize next. Even if the agent can directly operate visualization software, the state needed for visualization remains primarily in its current work environment. If the user, another agent, or a specialized visualization tool takes over, that state must first be organized, exported, or restated.

This handoff requires the work state produced during one stage to be directly usable in the next stage. By default, however, human and AI executors do not share a common work state. Regardless of who carries it out, task execution is always mediated by a work environment. A work environment maps the portion of the physical or virtual world relevant to a task into forms that an executor can perceive and manipulate. These representations constitute the \textit{work state}. Through the interface provided by the work environment, an executor can read and write the work state and invoke the capabilities needed to carry the task forward. Traditional operating systems and existing agent systems serve as work environments for their respective executors. Whether a handoff between humans and AI can be seamless therefore depends on whether their work environments share the state required to continue the task.

Existing agent systems provide separate work environments for human and AI executors. These environments commonly maintain separate, unshared work states. Consequently, when an executor changes, the task cannot continue seamlessly without an additional bridge. This problem has two causes. First, humans primarily carry work forward through graphical interfaces, whereas AI relies on the conversational context and memory in its own work environment. Their work states are therefore retained in different forms, and the state created in one phase cannot ordinarily become the basis for work in the next. Second, the work state available to AI is often limited by the interface provided by its current work environment. Even when AI is embedded in an application used by a human, it receives only a narrow interface exposed by that application, not the work environment occupied by the human. Such bridging generally takes one of two forms: developers build specialized interfaces that expose selected parts of the state from another work environment to an agent, or users manually migrate context when switching. The former cannot cover new scenarios as they arise; the latter repeatedly imposes alignment costs on users. As a task acquires more stages, the bridging burden grows and scalability declines.

To address this problem, we propose \textit{Human-AI Co-inhabitation}, a type of work environment that enables humans and AI to seamlessly take turns continuing work on the same task. This type of work environment requires humans and AI to share representations of the data and objects that constitute the work state, together with a unified, auditable interface. These conditions allow work to continue seamlessly when the executor changes, without scenario-specific bridges. To realize this goal, we derive three design principles. Work data should be represented as natural-language-readable text objects with stable structure that humans and AI can read and write directly, without customized access methods for each scenario. The work objects that a human is currently manipulating should have a unified, referable representation, so that AI can continue work without reidentifying them from screenshots or descriptions. System capabilities should be available to every executor through the same auditable entry point, eliminating private tool layers built specifically for AI. 

Based on these principles, we design and implement CrabOS, an operating system for a work environment shared by humans and AI. At the system layer, CrabOS organizes the data and objects that constitute the work state, along with a unified interface. Our case studies show that its operating-system primitives can be composed into capabilities for memory management, cross-application context continuation, and multi-agent task orchestration. CrabOS supports multiple modes of seamless task continuation between humans and AI, as well as among AI agents. It shifts the basis for supporting complex tasks with alternating leadership from application-level bridges to the operating-system layer.

%% file: sections/02_related_work.tex
\section{Related Work}
\label{sec:related-work}

Existing work has investigated the work environments that agent systems provide to their executors at multiple levels, including application integration, runtime organization, and system-layer abstraction. It broadly falls into three design paradigms. The first two each provide a work environment for only one kind of executor. The first centers on humans, embedding AI as a capability that assists people in advancing work. The second centers on AI, independently organizing the agent loop, sessions, permissions, and tool invocation above a host operating system to support autonomous progress on multistep tasks. A third paradigm attempts to place the agent runtime or execution environment directly at the system layer and reorganize agent capabilities with OS-like abstractions.

\subsection*{Human-Centric AI Assistance}

These agent systems treat AI as contextual assistance within an existing application, rather than as an independent executor that continues a task. GitHub Copilot \cite{github2026copilot} embeds code suggestions and conversational question answering in the development workflow, while the developer remains in control. OpenAI Prism \cite{openai2026prism} incorporates project-aware proofreading, citation retrieval, and formatting into a LaTeX scientific-writing workspace. Figma AI \cite{figma2026figmaai} offers a set of AI tools for the design process, and Adobe Firefly \cite{adobe2026firefly} places generation, editing, and refinement capabilities within design and creative workflows. Users continuously select, adopt, and revise outputs in the existing canvas and editing interface. These systems emphasize augmenting human work in established applications rather than allowing AI to run autonomously over shared system state for extended periods. CrabOS differs not in whether it offers contextual assistance, but in whether AI remains confined to a functional module within an application.

\subsection*{Autonomous Agent Systems}

Another class of agent systems organizes AI as a continuously running executor above the host operating system, with a runtime that maintains sessions, permissions, tool invocation, and multistep task progress. OpenClaw Gateway \cite{openclaw2026openclaw} uses a unified gateway to organize channels, sessions, and routing. Hermes Agent \cite{nousresearch2026hermesagent} combines long-running sessions, SQLite state, and multichannel interaction into a persistent agent runtime. Claude Code \cite{anthropic2026claudecode}, Codex CLI \cite{openai2026codex}, and OpenCode \cite{anomaly2026opencode} target local coding workflows, but respectively organize their agent loops around explicit session recovery, approval policies, and sandbox or permission rules. Manus \cite{manus2026manus} advances multistep tasks in an independent sandbox or cloud virtual machine and allows users to take over the browser or local computer when needed. These systems have developed mature forms of autonomous execution, permission gating, and task orchestration. Their state, capability boundaries, and interaction interfaces, however, usually remain encapsulated within their respective runtimes. In CrabOS, these concerns no longer stop at the runtime layer but are further moved down to the system layer.

\subsection*{Agent Operating Systems}

Other work directly considers moving the runtime, execution plane, or collaboration boundary of AI agents down to the system layer. AIOS \cite{mei2025aios} explicitly divides the agent runtime into kernel and SDK layers and describes its system abstractions through system calls, a scheduler, and managers for context, memory, storage, tools, and access. UFO2 \cite{zhang2025ufo2} targets Windows desktop automation and combines HostAgent, AppAgent, a unified GUI-API action layer, and a PiP isolated desktop as a desktop AgentOS. AOHP \cite{zhao2026aohp} further moves the agent harness into the Android system layer, treats agents as first-class OS actors, and restructures system and framework layers around personalized service composition, cross-application execution, and information-flow security. AgentOS from \texttt{agentos.to} \cite{agentos2026humanai} organizes local human-AI collaboration around an engine as the sole broker and a local memex graph as shared state, while retaining the GUI as an optional layer. In comparison, Rivet's agentOS \cite{rivet2026agentos} is closer to an in-process agent runtime, and the Agent OS in the Microsoft Agent Governance Toolkit \cite{microsoft2026agentgovernance} is closer to a governance kernel and policy engine. CrabOS also places AI agents at the system layer, but with a different focus. These agent operating systems primarily consider how agents can gain a stronger execution plane and finer-grained governance within a host system. Starting from Human-AI Co-inhabitation, CrabOS requires the work environment to make work data, work objects, and access interfaces available at the system layer to both human and AI executors.

%% file: sections/03_design_principles.tex
\section{Design Principles}
\label{sec:design-principles}

Human-AI Co-inhabitation aims to let human and AI executors continually continue each other's work on the same task in the same work environment. A work environment consists of work state and access interfaces. Work state, in turn, includes work data and work objects. Work data is content produced while a task advances and requiring persistent storage. Work objects are the concrete objects that an executor currently refers to or manipulates. They include persisted entities such as files, tasks, and pages, as well as active operational state such as selections and focus that has not yet been persisted as data. For humans and AI to share one work environment, three conditions must hold simultaneously: both must access work data uniformly, both must refer to work objects, and the access interface must be uniformly available to both. The three conditions correspond to costs borne by three roles. Developers need a low-cost way to add collaboration capabilities for new scenarios; users need to hand work objects to AI without restating them; and AI needs to invoke system capabilities without limits imposed by an agent system's implementation. Existing agent systems meet these needs only within their own boundaries and cannot organize work data, work objects, and access interfaces into structures shared across agent systems. Human-AI Co-inhabitation therefore follows three design principles.

\subsection*{Principle 1: Textualize Work Data}

In a single-executor setting, work data serves only that executor. Whether it is stored in a database, memory, or a private format does not impede progress as long as the executor can continue to use it, and developers need only design for that one access method. In Human-AI Co-inhabitation, the same work data must support a growing set of collaboration scenarios. Cross-session continuation requires importable and exportable data; multi-agent orchestration requires data that can be scheduled and composed; memory retrieval requires data that can be indexed and recalled. If the data remains sealed in private storage accessible only through specialized APIs, developers must design another dedicated access method whenever a collaboration scenario is added. As scenarios multiply, so do the dedicated access methods that must be maintained, repeatedly consuming development effort on the same class of problem.

Human-AI Co-inhabitation therefore requires work data to persist in a form that both humans and AI can directly process, without a custom access method for each scenario. Humans should be able to understand and modify the data directly, and AI should be able to parse and continue using it directly. Both should share one representation rather than access it through separate methods. Natural-language-based text objects meet this need. When task goals, progress, and intermediate results exist as natural-language text objects, reading, writing, and subscribing are ordinary file-system operations, requiring no per-scenario access mechanism. This yields the first principle: work data must be represented as natural-language-readable text objects with stable structure.

\subsection*{Principle 2: Share Work Objects}

In a single-executor setting, the current work object need only serve that executor's continuous operation. Through a graphical interface, a user can tell which file is open, what content is selected, and which window has focus. Even if this information exists only as private interface state within an application, it does not impede the user from continuing work. Purely autonomous AI faces the same absence of a problem because AI does not need to continue from the interface currently viewed by a human. It only needs the state and capabilities required by its task. The challenge in Human-AI Co-inhabitation is that when a human says, ``continue editing this page,'' the page is clear to the human but is not represented by the system as an object that AI can directly refer to. AI is not unable to see the screen. It can obtain an image through a screenshot, but it must still reidentify the page, selection, and focus before it can continue. Screenshots, copy and paste, and verbal restatement therefore become bridging costs paid at every handoff. These work objects already exist in the system but cannot be directly reused by the new executor.

Human-AI Co-inhabitation therefore requires humans and AI to directly manipulate the same set of referable work objects. Work objects include not only persisted entities such as files, tasks, and pages, but also active content in operation, such as the current window, selection, and focus position. Only when these work objects have unified, referable representations can the content that a human is viewing, selecting, or editing directly become usable work objects when AI takes over, and can AI's results be directly continued by the human on the same work objects. This yields the second principle: humans and AI must directly manipulate the same set of referable work objects.

\subsection*{Principle 3: A Unified Capability Entry}

In a single-executor setting, users can reach nearly all system capabilities directly through graphical interfaces, the file system, or a terminal, without a special intermediary opening permission for each access. Independently deployed AI also presents no fundamental problem: developers can implement the tools needed for one specific task, and a narrower task boundary requires fewer tools. In Human-AI Co-inhabitation, AI shares a work environment with humans. The relevant scope is therefore the full range of capabilities that humans can reach through graphical interfaces, rather than a narrowly predefined task. What AI can invoke depends on whether an agent system has deliberately packaged that capability as a tool. If it has not, AI cannot use the capability even when the system already provides it. This private packaging layer continuously limits the efficiency of AI-system interaction, leaving AI with only a finite set of capabilities enumerated in advance.

Opening every system capability directly to AI is not feasible. When humans and AI share a work environment, AI behavior cannot be presumed to have the same trust level as human action. A human's request to write the document currently being edited may be admitted directly, while a broad modification initiated by AI may require additional authorization. The requirements to expose capabilities and control risk can be satisfied only when a capability is invoked. The system must know who requests which capability in order to impose different authorization and constraints. At the system layer, an executor is thus an identifiable calling subject. In Human-AI Co-inhabitation, AI should not depend on an agent system's private tool-packaging layer to reach system capabilities. Like a human, it should invoke capabilities through the same unified, auditable entry, allowing the system to identify the calling subject, resolve the target, and enforce policies at invocation time. AI can then use capabilities already provided by the system without waiting for an agent system to package them as tools. This yields the third principle: eliminate private tool-packaging layers for AI, and require every executor to invoke system capabilities through the same auditable entry.

%% file: sections/04_system_design.tex
\section{System Design}
\label{sec:system-design}

\begin{figure}[t]
\centering
\includegraphics[width=\textwidth]{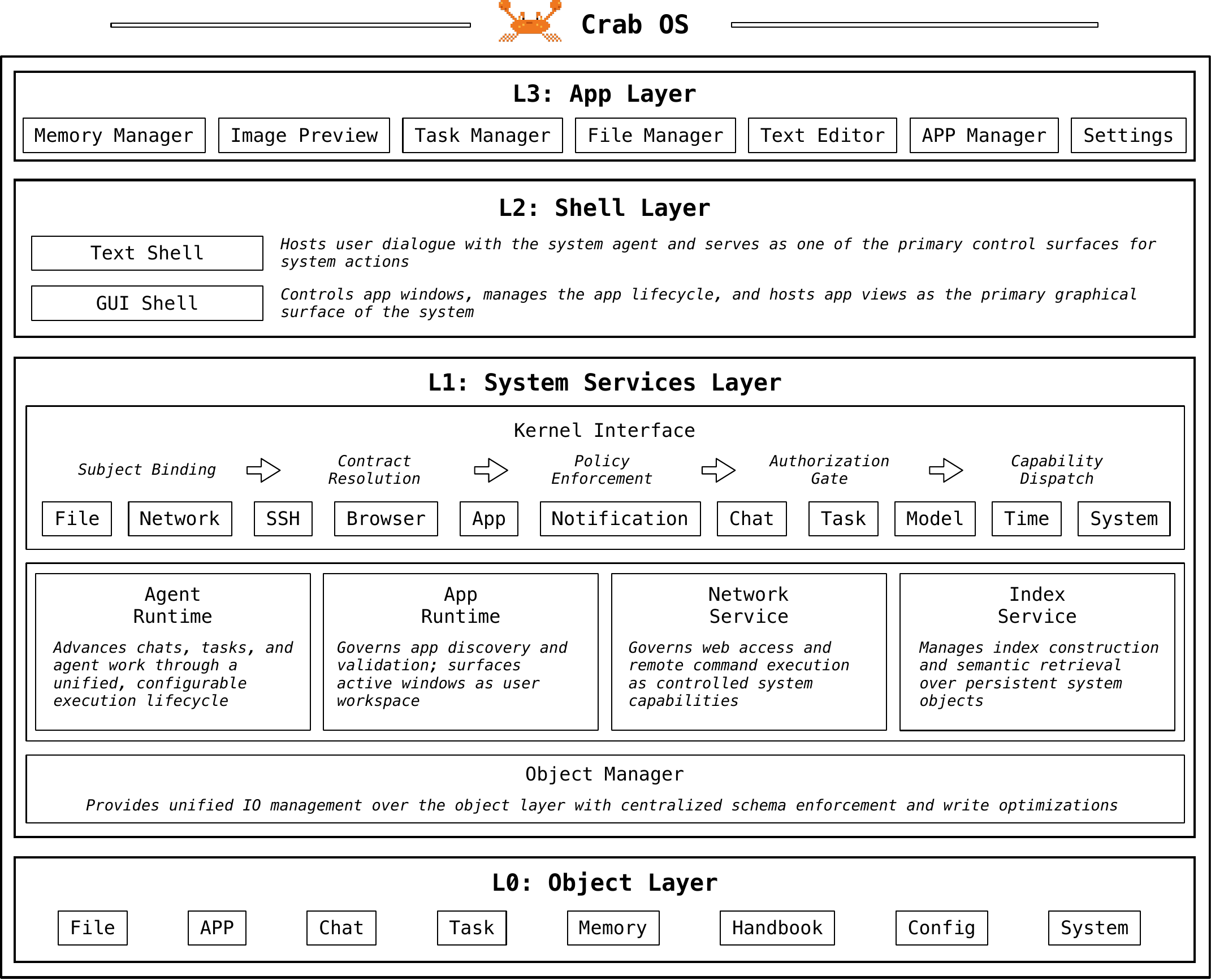}
\caption{The four-layer architecture of CrabOS and the responsibility of each layer.}
\label{fig:crabos-main}
\end{figure}

CrabOS has four layers: the L0 Object Layer, the L1 System Services Layer, the L2 Shell Layer, and the L3 App Layer. L0 persists all state. L1 uniformly schedules and executes system capabilities. L2 manages interaction interfaces for humans and applications while also exposing interface state to AI. L3 extends business functionality through apps. Figure~\ref{fig:crabos-main} shows the system architecture.

\subsection*{L0 Object Layer}

A database is the most intuitive persistence choice for an agent system, and most agent systems do use one. Its less apparent cost is that, once state enters the database, the sole legitimate path to that state is an API. Supporting a hook requires designing a hook API; supporting sub-agents requires an orchestration API; sharing context across AI agents requires a message-passing API. The system can do only what was enumerated in advance, and a capability that developers have not implemented does not exist.

CrabOS therefore does not encapsulate state in storage accessible only through a query protocol. Instead, it materializes every system object as a text object and uses memory only for caching. The file system is chosen because it provides natural direct addresses and general-purpose I/O for text objects; the file wrapper itself is not the design principle. Object content is the LLM-readable state representation, not an internal format awaiting decoding by a specialized client. These objects are not unstructured text. They are state units with stable structure that can be uniformly validated, indexed, and processed incrementally. This design yields properties unavailable from database-based designs: many capabilities need not be designed at the system layer, but emerge directly from addressable text objects. A hook is a subscription to changes at a path; a sub-AI agent is the creation of a Task object followed by waiting for completion; sharing context across AI agents is reading another agent's Memory object. The required primitives, \texttt{WaitPathChange}, \texttt{ReadFile}, and \texttt{WriteFile}, already exist as tools in the Kernel Interface. Orchestration patterns thus require no separate kernel implementation and can be composed by AI agents from these primitives.

The L0 Object Layer has eight object types, and removing any one breaks an access contract. Humans write \textbf{File} for humans to read and \textbf{Handbook} for AI agents to read. Before every reasoning step, CrabOS proactively injects the Handbook directory into context, so merging them would eliminate the injection boundary. AI execution writes three object types: \textbf{Chat}, \textbf{Task}, and \textbf{Memory}. Chat is interaction-driven and waits for a human message at each step; Task executes autonomously after entering the queue until it completes; Memory distills information across sessions and has a scope beyond a single interaction. Their indexing strategies, execution models, and permission boundaries differ accordingly. CrabOS writes \textbf{App}, \textbf{Config}, and \textbf{System}. App stores installation information, Config is the authoritative source of behavior, and System is secondary data derived at runtime. Config must exist for CrabOS to run. System can be deleted and automatically rebuilt because it is never authoritative. System also cannot be renamed Log: a log is an append-only event stream, whereas System contains retrieval indexes, pending notifications, and browser state that is overwritten in real time, each with different semantics.

\subsection*{L1 System Services Layer}

The L1 System Services Layer has three structural tiers. At the top, the Kernel Interface is the unified entry for every external invocation. The middle tier contains services. At the bottom, the Object Manager writes the state of all services to the L0 Object Layer that it manages. Invocations flow from top to bottom. Any subject, whether an AI agent, an app, or a user, must pass through the complete Kernel Interface pipeline before a system capability reaches the service that executes it.

The \textbf{Kernel Interface} has five mandatory stages. \textbf{Subject Binding} binds every invocation to an identified system subject, whose identity is enforced by the system at the transport layer rather than self-declared. \textbf{Contract Resolution} interprets a request as a governed system contract and establishes the type and boundary of the capability. \textbf{Policy Enforcement} determines permission and resource boundaries before execution and rejects an invocation that exceeds them. \textbf{Authorization Gate} suspends an operation requiring explicit user authorization under a security policy until confirmation arrives. \textbf{Capability Dispatch} sends an invocation that has passed all checks to the service responsible for it.

At the middle tier, the \textbf{Agent Runtime} advances chats, tasks, and agent work through their respective execution lifecycles. It provides two modes: Chat interacts with humans asynchronously through text, whereas Task runs autonomously to completion after receiving a goal. An AI agent's behavior is determined by a template comprising a system prompt, state-information slots, and an available tool set. Different interaction forms are therefore different template configurations, not orchestration patterns preordained by the kernel. Claude Code's plan mode \cite{anthropic2026claudecode} and Codex's read-only mode \cite{openai2026codex} can both be expressed through this mechanism. The \textbf{App Runtime} manages app discovery and verification, and presents the focused window as the user workspace so the system can perceive the user's current view. The \textbf{Index Service} builds and queries indexes over the full object space. The \textbf{Network Service} uniformly manages web browsing, external network access, and remote command execution as controlled system capabilities.

The \textbf{Object Manager} underpins the L1 System Services Layer. It provides unified object I/O to every service and centrally performs schema validation and write optimization. Every L1 service reads and writes L0 objects through the Object Manager, ensuring consistency and write performance in this layer.

\begin{figure}[t]
\centering
\includegraphics[width=\textwidth]{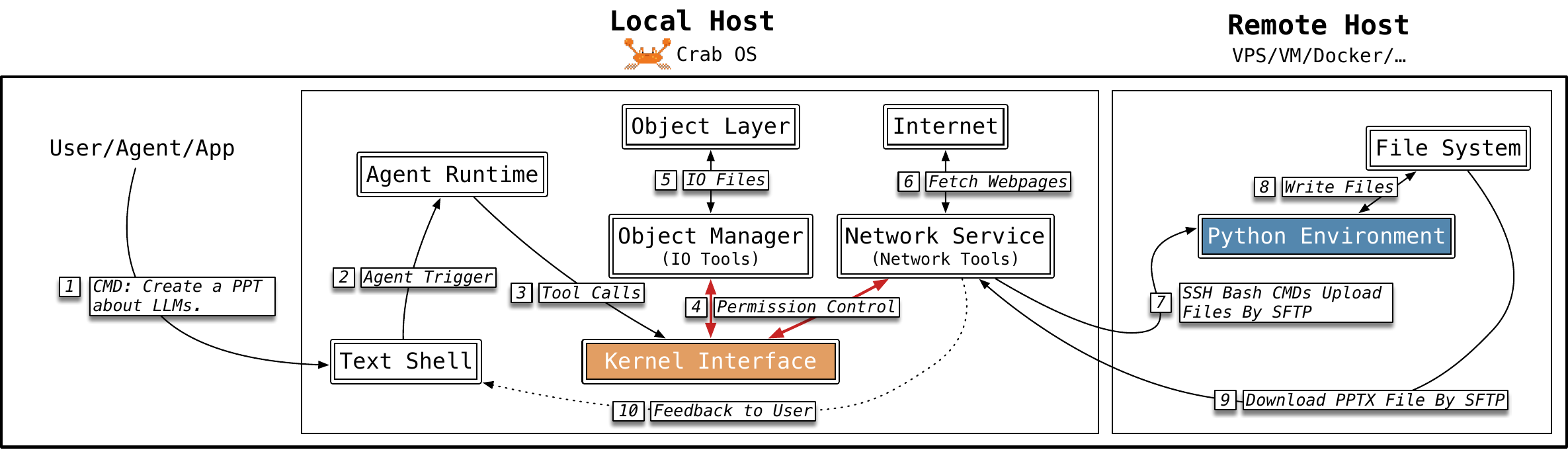}
\caption{The invocation path of an actual task in CrabOS.}
\label{fig:crabos-case}
\end{figure}

Figure~\ref{fig:crabos-case} presents the complete invocation path for an actual task in the L1 System Services Layer. After a user issues an instruction in the Text Shell, the Agent Runtime starts an AI agent to execute the task. Every tool invocation generated during execution first passes through the five-stage Kernel Interface pipeline for subject binding, contract resolution, and permission adjudication before dispatch to a concrete service. The illustrated task involves both web retrieval and code execution. The Network Service performs web fetching, and intermediate data and results are written to the L0 Object Layer through the Object Manager. Code execution likewise reaches an external execution environment through the Network Service. The environment may be a VPS, virtual machine, container, or isolated runtime on the same machine. CrabOS makes no assumption about its internal structure and treats it solely as an execution endpoint accessed through the network. Whether an AI agent reads local files, fetches web pages, or runs code in an external environment, the same system-level path applies: the capability invocation enters through the Kernel Interface, the system identifies the subject and adjudicates permissions, a service performs the invocation, and all resulting state is ultimately stored in the L0 Object Layer.

\subsection*{L2 Shell Layer}

In CrabOS, the L2 Shell Layer is not only a human operation interface, but also the layer through which AI perceives human work state. The currently running apps, the focused window, and the content in that window have one authoritative source: the entity that manages every app window's state. Principle 2 requires this entity to be open to both humans and AI. It cannot reside in L3, where it would be at the same level as the apps it manages and violate layering. It also cannot reside in L1, where a headless computation layer would have to render a UI and its responsibilities would become confused. It can only be an independent layer between L1 and L3: the L2 Shell Layer that manages every L3 app window.

CrabOS adopts a deliberately restrained design: the smallest built-in surface and the greatest room for extension. Every function that can become an app should become one. File management, search, settings, and even the app manager itself are all implemented as apps. Exactly two L2 modules are irreplaceable. The \textbf{GUI Shell} contains the Dock and manages app launching and app window hosting. The \textbf{Text Shell} is the terminal through which users converse with system AI agents. Like a traditional operating system's command line, it must exist before all apps. Otherwise, the system would be silent to users until an app loaded. Every app runs in an isolated environment, and every capability invocation travels through the Kernel Interface on the same unified path as AI-agent invocations. Apps run without users configuring any service and are usable as soon as they are generated.

\subsection*{L3 App Layer}

The L3 App Layer is where CrabOS differs most from existing agent systems. Existing systems implicitly assume that people submit requests only through a conversational interface and receive textual task results. Some systems provide generative interfaces for charts and other content, but such interfaces remain secondary presentations of agent-system outputs whose capability boundaries are predefined by that system. In CrabOS, apps are peers of system AI agents. Through the SDK, an app has the same system API invocation capabilities as an AI agent, and can therefore implement agent logic itself and become an AI agent.

CrabOS preinstalls a core set of apps: Memory Manager, Image Preview, Task Manager, File Manager, Text Editor, App Manager, and Settings. They cover common operating-system functions, but all are L3 apps. The L2 Shell Layer contains no built-in business logic, and every one of these apps can be freely uninstalled. CrabOS consistently follows the rule that anything capable of becoming an app should become one. Even the file manager and system settings are not exceptions, and no app is hard-coded into the system.

%% file: sections/05_implementation.tex
\section{Implementation}
\label{sec:implementation}

CrabOS is built on Electron \cite{electron2026electron}, with web technologies as its implementation foundation. This choice gives CrabOS native cross-platform desktop support and lets the L2 GUI Shell and L3 apps share one rendering environment, removing the need to bridge a native UI layer and a web layer.

L3 apps use iframes as their runtime, relying on a browser security model tested over decades. Browsers have limited file-I/O capabilities, and the app SDK provided by L1 fills this gap. Apps use the SDK to invoke system APIs through the same unified entry as AI agents, preserving capability boundaries and security control. An iframe also supports WebAssembly and WebGPU, so computationally intensive work can run directly inside it. When an app needs system capabilities, it still invokes them through the SDK and cannot bypass the Kernel Interface.

Existing native apps can enter CrabOS through three paths of increasing integration depth. The first is to implement a complete L3 app from scratch, with both interface and logic inside the iframe and SDK. The second is to rewrite only the L3 GUI while retaining the original native app's API as the backend, which runs in a sandbox. The third is a GUI adapter. It likewise first implements an L3 interface for human and AI operation, while the native app continues to run on the host or remotely. Operations in the L3 interface are first mapped to UI controls, then to the corresponding control coordinates, which drive the native app. To humans and AI, the entry is the same L3 app; only the way its backend reaches the original capability differs.

CrabOS is designed to work out of the box. Users need not configure external services after installation, and core features must fully run with zero dependencies. The Index Service therefore performs embedding locally. CrabOS loads ONNX \cite{onnx2026onnx} embedding models through Transformers.js \cite{huggingface2026transformersjs} and performs inference directly in the Node.js process, without relying on a cloud embedding API. Its vector index works fully offline and is unaffected by changes in third-party services.

%% file: sections/06_case_studies.tex
\section{Case Studies}
\label{sec:case-studies}

This section starts from concrete capabilities and compares the implementation paths that CrabOS and mainstream agent systems take for the same capability, showing how architectural differences manifest in practice.

\subsection*{Memory Management}

Memory management is one of the areas in which current agent systems diverge most sharply. OpenClaw \cite{openclaw2026openclaw} uses two Markdown-file layers for long-term memory and daily logs, and triggers a memory flush as the context window approaches compression limits. Hermes \cite{nousresearch2026hermesagent} offers a built-in two-file mechanism together with eight pluggable external memory providers, each with a different storage backend and retrieval algorithm. These approaches make different tradeoffs but share one characteristic: their memory mechanism is built into, or tightly coupled to, the agent system. Switching requires either changing agent systems or migrating data.

In CrabOS, any memory strategy can be independently implemented as an L3 app. The system uses the memory mechanism implemented by the app a user installs. Hermes's provider mechanism may appear flexible, but it only selects among predefined options with integration interfaces fixed by Hermes. CrabOS makes flexibility available at a lower layer. Chat and Task objects in L0 are text objects that any app can directly read; written-back content is automatically injected into reasoning through the slot mechanism; and the Index Service supplies full-text and vector retrieval. Anyone can build on these primitives to implement OpenClaw-style active AI memory writes, Hermes-style post-session distillation, synchronization with an enterprise external vector database, or a fully custom knowledge-graph structure. These implementations are independent: they can coexist or be replaced at any time, and require no knowledge of CrabOS internals beyond reading and writing these objects.

\subsection*{Vibe Anything}

After ``vibe coding'' became popular, vibe writing, vibe reading, and vibe anything became common expectations for AI collaboration. Users expect AI to know where they are writing and which page they are reading, so collaboration can start at any time without reconstructing context. Current practice is to subscribe to separate AI agent systems. AI writing assistants, AI reading tools, and AI design collaborators each serve their own scenario and implement their own kind of ``vibe.'' Even within one agent system, the supported type of vibe depends on the UI it includes, and its coverage is determined by developer priorities.

In CrabOS, every scenario expressed as an app can have this capability. A PDF viewer is an L3 app whose runtime state is an L0 object. When users turn a page, the state is immediately written to an addressable object that an AI agent can directly read, without awaiting any transfer. Document editing works the same way: a document is a directly understandable text object in L0, and the human and AI manipulate the same state. The AI agent need not wait for an app to translate editing context into a separate interface. In CrabOS, this capability comes from system architecture rather than a specific implementation by an application. It covers all scenarios represented as apps, allowing CrabOS to provide ``vibe anything.''

\subsection*{Multi-Agent Task Orchestration}

Task orchestration is among the fastest-moving capabilities in current agent systems. Codex \cite{openai2026codex} provides a \texttt{/goal} mode: a user sets a goal and stopping condition, and the system then runs autonomously for hours until that condition is met or its budget is exhausted. Hermes \cite{nousresearch2026hermesagent} centers on a SQLite task board whose dispatcher automatically claims tasks and starts corresponding specialist agents, with dependency graphs, crash recovery, and complete execution history. OpenClaw \cite{openclaw2026openclaw} uses its Lobster engine to replace LLM planning with YAML-configured deterministic workflows, turning multi-agent orchestration from temporary model decisions into controllable declarative pipelines. Claude Tag \cite{anthropic2026claudetag} introduces an ambient mode in which AI agents persist as Slack-channel members, accumulate context, and may proactively address unresolved tasks without an explicit \texttt{@}. These features emphasize different concerns but share an underlying structure: work data is private to its agent system, and external access to subtasks, progress, and execution history is limited to that system's UI or API. Each additional orchestration style requires another mechanism over private state.

In CrabOS, Task is an L0 text object. Every orchestration style can therefore be composed from existing object primitives rather than designed as a new mechanism. Goal mode is a task loop that continues until its completion condition is satisfied. A Kanban board is an app that subscribes to and presents Task state. Hooks and callbacks are subscriptions to changes in Task paths through \texttt{WaitPathChange}. Ambient behavior is an AI agent that subscribes to L0 state changes and proactively advances work without an explicit trigger. These capabilities do not require separate system-layer implementation. They naturally emerge because objects are addressable and their content is directly understandable.

\subsection*{Operation Traces}

In human-centric AI assistance systems, AI can usually see changes made after a human continues a task. An editor provides a diff to Copilot, or a design tool provides layer changes to its embedded AI. This depends on each application developing scenario-specific logic for perception and recording. The applications covered and the granularity of observable changes are separately determined by their developers and cannot be reused across applications.

In CrabOS, capability invocations from every L3 app pass through tool gating in the Kernel Interface. Subject Binding and Capability Dispatch already bind the calling subject, capability, and parameters to every invocation, whether it originates from a human or AI. Persisting these invocations naturally produces a trace at the granularity of one semantic operation, without a recording module designed specifically for continuation. When an AI agent takes over, it can read the trace to learn what the human recently did without being told. This trace data can also support post-training of large language models.

%% file: sections/07_discussion.tex
\section{Discussion}
\label{sec:discussion}

CrabOS makes several explicit tradeoffs. They define both the scenarios for which it is suited and the boundaries it does not yet cover.

\textbf{Bash ecosystem:}
Many agent systems use Bash as a general execution interface, allowing AI agents to run system commands and established toolchains directly. CrabOS provides no built-in arbitrary Bash execution because any shell execution could bypass the Kernel Interface and violate the system model centered on a unified entry. Consequently, CrabOS has more limited native support for Bash than some agent systems. To address this limitation, CrabOS uses SSH to interact with an external execution environment, such as a local sandbox, container, virtual machine, or remote host, where it runs Bash or other commands. CrabOS only initiates and manages this controlled invocation, while the user chooses the form of the execution environment and CrabOS makes no assumption about its internals. This approach uses the existing shell-tool ecosystem without violating the system model based on a unified entry.

\textbf{Boundary of the CrabOS ecosystem:}
Principle 2 requires humans and AI to manipulate the same referable work objects, which is why L3 provides a shared iframe interface. This means that an existing native application must first be migrated into an L3 app before it can participate in these work objects. Migration does not require a complete rewrite. The GUI-adapter approach described above leaves the native app backend unchanged, adds only an L3 interface, and maps UI interactions to the original native app. Humans and AI can thus still share most work objects in CrabOS. This approach also records real human operation traces, which can improve human-AI task continuation and provide data for post-training large language models.

\textbf{I/O performance of persisted text objects:}
Persisting the complete L0 Object Layer introduces more write latency than an in-memory database. AI-agent tasks, however, do not have I/O-intensive patterns. Conversations, task state, and memory entries are written far less frequently than in typical database workloads, so this latency is acceptable. For high-frequency writes such as streaming output, CrabOS uses delayed writes: content is first buffered in memory, then persisted in batches after a threshold or interval. This preserves the streaming experience.

%% file: sections/08_conclusion.tex
\section{Conclusion}
\label{sec:conclusion}

Complex tasks often require humans and AI to take turns moving a task forward, each handing it over to the other after completing a stage. Whether such a handoff can be seamless depends on whether the work state produced during one stage can be directly used in the next. We propose Human-AI Co-inhabitation as a type of work environment that requires humans and AI to share representations of the data and objects that constitute the work state, together with a unified, auditable interface. It reduces the collaboration costs borne by developers, users, and AI. Existing agent systems do not provide work environments that meet these conditions. The data and objects that constitute the work state, as well as the interfaces through which they are accessed, are confined within the separate boundaries of their respective work environments, so seamless continuation depends on additional bridges: specialized interfaces or manual migration. As task stages multiply, the bridging burden increases.

This paper presents CrabOS, an operating system designed to realize Human-AI Co-inhabitation. CrabOS follows three principles: it persists work data as natural-language-readable text objects with stable structure; it lets humans and AI directly manipulate the same set of referable work objects; and it eliminates private tool layers built specifically for AI by requiring every executor to invoke system capabilities through the same auditable entry point. These principles reduce collaboration costs for developers, users, and AI simultaneously. Typical capabilities, including agent memory management, seamless cross-application context continuation, and multi-agent task orchestration, can be composed from operating-system primitives without additional bridges. As AI becomes a long-running executor, we believe operating systems built on the three principles of Human-AI Co-inhabitation will become infrastructure for future agent systems.